\documentclass[letterpaper]{article}

\usepackage{natbib,alifeconf}  
\usepackage{amsmath,amssymb,amsfonts}
\usepackage{algorithm,algorithmic}
\usepackage{enumitem,subcaption,xcolor,microtype}
\usepackage{url,hyperref,cleveref}
\usepackage{booktabs}
\usepackage{stfloats}  

\usepackage{lipsum}

\newcommand\blfootnote[1]{%
  \begingroup
  \renewcommand\thefootnote{}\footnote{#1}%
  \addtocounter{footnote}{-1}%
  \endgroup
}

\title{Multi-Behavioral Evolved Substrates Through\\Neuromodulation and Activation Selection}

\author{
    Romain Claret$^{1}$,
    Michael O'Neill$^{2}$,
    Paul Cotofrei$^1$, \and
    Kilian Stoffel$^1$ \\
    \mbox{}\\
    $^1$University of Neuch\^atel, Switzerland \\
    $^2$University College Dublin, Ireland
    romain.claret@unine.ch
} 

\begin{document}

\maketitle

\begin{abstract}
Open-ended artificial life systems must acquire diverse competencies from a single evolving genotype. Biological brains combine neuromodulation, which reconfigures circuits without changing connections, with diverse neuron types matched to specific computational roles. Can artificial evolution achieve something analogous in indirectly encoded substrates?

Using indirectly encoded substrates evolved via CPPNs, we show through more than 10,000 experiments that neuromodulation alone is insufficient: under evolutionary search, monotonic activation functions impose a 75\% ceiling on parity tasks that persists regardless of capacity, topology, or population size. This is an evolutionary search barrier, not a representational limit, since Adam gradient descent achieves 100\% on the identical architecture.

We combine neuromodulation with per-task activation function selection, matching oscillatory primitives to parity tasks and monotonic to threshold tasks, producing multi-behavioral evolved substrates. The result: 100\% simultaneous 5-task success across all 30 seeds (median 14 generations). This generalizes across the oscillatory activation class: all four functions reach 100\% (30 seeds each). Neither mechanism suffices alone.

The barrier extends to higher-arity and asymmetric tasks, while multi-layer depth provides an alternative path. For open-ended evolution, the computational primitive should itself be an evolvable trait. At inference, one evolved genotype expresses many behaviors.
\end{abstract}

Data/Code available at: \url{https://github.com/RomainClaret/emr-hyperneat}
\blfootnote{\textcopyright{} 2026 Romain Claret, Michael O'Neill, Paul Cotofrei, Kilian Stoffel. Published under a Creative Commons Attribution 4.0 International (CC BY 4.0) license.}

\section{Introduction}

Catastrophic interference~\citep{mccloskey1989catastrophic,french1999catastrophic} remains a central obstacle for evolving systems that must acquire multiple competencies. Artificial neuromodulation has shown promise for continual learning~\citep{ellefsen2015neural,miconi2018differentiable}, drawing on biological evidence that chemical signals reconfigure neural circuits without changing synaptic connections~\citep{doya2002metalearning,marder2012neuromodulation}. Yet biological brains rely on a second mechanism that artificial systems largely ignore: neurons with distinct electrophysiological profiles matched to specific circuit roles~\citep{kepecs2014interneuron}, enabling different computational primitives within the same circuit. A fundamental question remains: is neuromodulation alone sufficient for simultaneous multi-task learning, or is computational primitive diversity also required?

That a single perceptron cannot compute XOR is a classical result~\citep{minsky1969perceptrons}. With a hidden layer, the representation exists, but what has \emph{not} been studied is whether evolutionary search can discover it when neuromodulation constrains the weight-receptor space. The single-task ceiling becomes an \emph{emergent multi-task barrier}, pair-specific, optimizer-dependent, and invisible to single-task analysis. This paper maps the barrier systematically across five Boolean tasks (XOR, AND, OR, NAND, NOR) in more than 10,000 experiments. The contribution is the systematic demonstration that activation choice and neuromodulation \emph{interact} to determine multi-task feasibility, beyond the known result that oscillatory activations solve parity~\citep{sitzmann2020siren,noel2021gcu}.

This question is especially relevant for indirectly encoded substrates, where a generative network, a Compositional Pattern Producing Network (CPPN), produces the substrate topology. Such substrates yield sparse, irregular structures that a human would be unlikely to design, yet evolution can discover weight-receptor configurations that, combined with neuromodulation, enable a single evolved topology to express qualitatively different behaviors across tasks. Unlike prior work on per-node activation function evolution~\citep{claret2026activations}, which gives each neuron its own activation function within a single task, the present approach gives all neurons the same activation but switches it between tasks, so the whole substrate changes computational mode depending on the task environment.

In the context of indirectly encoded substrates evolved via CPPNs, we address three research questions:
\pagebreak
\begin{description}
    \item[\textbf{RQ1}] Does activation function choice create systematic multi-task barriers under neuromodulation?
    \item[\textbf{RQ2}] Is the barrier representational or optimizer-dependent?
    \item[\textbf{RQ3}] Does per-task activation function selection resolve the barrier, and is neuromodulation still necessary?
\end{description}

\textbf{Contributions:}
\begin{enumerate}
    \item A systematic mapping of multi-task compatibility under neuromodulation, revealing that task compatibility depends on the match between activation functions and task computational structure, not neuromodulation parameters, topology, or population size (RQ1).
    \item Root cause identification: at a single layer, monotonic activations impose a 75\% ceiling on parity in multi-task evolutionary search, creating an optimizer-dependent barrier (gradient descent bypasses it) that oscillatory primitives or added depth can resolve (RQ2).
    \item A bio-inspired solution: per-task activation functions matched to computational structure enable a single evolved network to reconfigure its behavior at inference, achieving 100\% 5-task success (all 30 seeds) with all 2-task barriers eliminated (RQ3).
    \item Robustness of the mechanism across an extensive sweep: the oscillatory advantage generalizes over the activation class (all four functions), the barrier extends to higher-arity and asymmetric tasks, and multi-layer depth provides an alternative resolution path.
\end{enumerate}

\section{Related Work}
\label{sec:related}

\textbf{Continual learning.} Elastic Weight Consolidation (EWC)~\citep{kirkpatrick2017overcoming} and Synaptic Intelligence~\citep{zenke2017continual} protect important weights via regularization, addressing catastrophic forgetting but not activation function mismatch. Progressive Networks~\citep{rusu2016progressive} add frozen columns per task, avoiding interference but scaling poorly. PackNet~\citep{mallya2018packnet} uses iterative pruning for task-specific subnetworks. \citet{velez2017diffusion} showed diffusion-based neuromodulation eliminates forgetting (91.8\% vs.\ 29.6\% retention). We find that even successful neuromodulation cannot resolve fundamental task conflicts arising from activation function choice, a dimension these methods do not address.

\textbf{Neuromodulation in AI.} \citet{soltoggio2008evolutionary} introduced evolved modulatory neurons for adaptive behavior. \citet{ellefsen2015neural} combined modularity with neuromodulation, achieving reduced forgetting through structural separation. Backpropamine~\citep{miconi2019backpropamine} enabled differentiable neuromodulated plasticity. These works focus on modulating existing computations. We show that modulation alone cannot overcome activation-task mismatch under evolutionary search.

\textbf{Multi-task learning.} Hard parameter sharing~\citep{caruana1997multitask} assumes positive transfer, an assumption we challenge. Gradient-based multi-task methods such as Model-Agnostic Meta-Learning (MAML)~\citep{finn2017maml} and Projecting Conflicting Gradients (PCGrad)~\citep{yu2020pcgrad} address task interference through meta-learning and gradient projection. These operate in multi-layer gradient-trained networks where the single-layer barrier we identify does not arise. Mixture-of-Experts architectures~\citep{jacobs1991adaptive,shazeer2017outrageously} route inputs to specialized subnetworks. The per-task activation mechanism proposed here is analogous but operates at the computational primitive level. \citet{driscoll2024dynamical} identified shared dynamical motifs in multi-task RNNs, paralleling the finding here that shared intermediate representations under uniform activation determine task compatibility. A complementary line studies \emph{neural reuse}, where the same neural components support multiple behaviors~\citep{candadai2018multifunctionality,benson2020neuralreuse}; the present work instead keeps the network fixed across tasks and changes only the activation primitive and modulatory input.

\textbf{Oscillatory activations.} Sinusoidal Representation Networks (SIREN)~\citep{sitzmann2020siren} showed that periodic functions preserve high-frequency information. The Growing Cosine Unit (GCU)~\citep{noel2021gcu} showed oscillatory activations enable single-neuron XOR solutions. In the NeuroEvolution of Augmenting Topologies (NEAT) lineage, activation function diversity progressed from fixed sigmoid~\citep{stanley2002evolving} through random assignment in CPPNs~\citep{stanley2009hypercube} to mutable per-node activation~\citep{hagg2017evolving}. The per-task mechanism proposed here is orthogonal: rather than diversifying functions across nodes, we assign a single function to all nodes per task, enabling the same network to switch computational primitives at inference.

\section{Methods}
\label{sec:methods}

The approach evolves a single neural network whose topology and weights are shared across all tasks; it reconfigures its behavior per task via two mechanisms: neuromodulation (gain, bias, and gating control) and per-task activation function selection.

\subsection{Neuromodulation Mechanism}

Neuromodulation is modeled as a 4-dimensional vector $\mathbf{NT} = [\text{DA}, \text{5HT}, \text{NE}, \text{ACh}]$ that adjusts neuron sensitivity through gain modulation~\citep{salinas2000gain,ferguson2020gain} without changing learned weights. The four dimensions are labeled after biological neuromodulators (dopamine, serotonin, norepinephrine, acetylcholine) but carry no physiological meaning; their roles are defined by the equations below.

For a hidden layer with pre-activation values $\mathbf{h}$, define the per-neuron modulatory projection $\mathbf{p} = \mathbf{R}_{:,1{:}3}\, \mathbf{NT}_{1{:}3}$, where $\mathbf{R} \in \mathbb{R}^{N \times 4}$ is an evolved receptor density matrix\\ ($N$ hidden neurons; per-neuron sensitivity to each modulator). The neuromodulated output is:
\begin{equation}
    \mathbf{g}_{\text{eff}} = \mathbf{g}_{\text{base}} + s \cdot \mathbf{p}
    \label{eq:gain}
\end{equation}
\begin{equation}
    \mathbf{h}_{\text{out}} = f(\mathbf{g}_{\text{eff}} \odot \mathbf{h} + \mathbf{b}_{\text{mod}}) \odot \boldsymbol{\gamma}
    \label{eq:forward}
\end{equation}
where $\mathbf{g}_{\text{base}}$ is a per-neuron base gain determined by the evolved substrate, $s = 5.0$ is a fixed scaling hyperparameter, $\mathbf{b}_{\text{mod}} = s \cdot \mathbf{p}$ shifts decision thresholds per task, $\boldsymbol{\gamma} = \sigma(\mathbf{p})$ provides sigmoid gating (enabling task-specific subnetwork selection), and $f(\cdot)$ is the activation function. Gain modulation, bias, and gating all derive from the single projection $\mathbf{p}$; they are coupled, not independently controlled.

The fourth component (ACh) controls output polarity via interpolation between the network output $y$ and its complement:
\begin{equation}
    y_{\text{final}} = \text{ACh} \cdot y + (1 - \text{ACh}) \cdot (1 - y)
    \label{eq:ach}
\end{equation}
where $y = \sigma(\mathbf{h}_{\text{out}}^\top \mathbf{w}_{\text{out}})$ is the sigmoid output. In all experiments ACh $\in \{0,1\}$, so Eq.~\ref{eq:ach} reduces to identity (ACh$=$1.0) or inversion (ACh$=$0.0); inversion enables NAND/NOR from AND/OR weights without changing the hidden-layer computation.

\textbf{Task-specific NT presets.} We use fixed NT presets: XOR $[0.95, 0.05, 0.95]$, AND $[0.10, 0.90, 0.10]$, OR $[0.50, 0.50, 0.50]$. These values were hand-designed to separate the three threshold tasks in modulatory space rather than obtained by parameter search. A \emph{compositional} design exploits NAND$=$NOT(AND) and NOR$=$NOT(OR) by reusing AND and OR profiles with ACh$=$0.0 (their geometry shapes task compatibility, analyzed in \nameref{sec:compatibility}).

\begin{figure*}[!b]
\centering
\includegraphics[width=\textwidth]{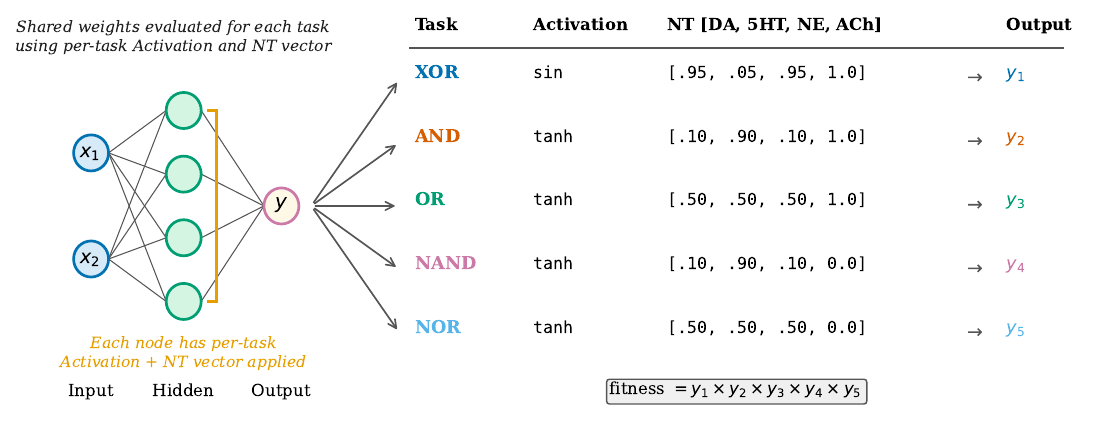}
\caption{One genotype, five phenotypic behaviors. (Left) A single evolved network (2 inputs, $N$ hidden, 1 output) with shared weights. (Right) The same network is evaluated five times with task-specific NT vectors and activation functions. XOR uses sin; all threshold tasks use tanh. NAND and NOR reuse AND and OR profiles with output inversion (ACh$=$0.0).}
\label{fig:architecture}
\end{figure*}

\subsection{Network Architecture and Encoding}

We evolve neural networks using EMR-HyperNEAT (Eager Multi-Resolution HyperNEAT)~\citep{claret2026emr}, a tensor-accelerated reformulation of Evolvable-Substrate HyperNEAT (ES-HyperNEAT)~\citep{stanley2009hypercube,risi2012enhanced}. A CPPN~\citep{stanley2007cppn} is evolved by NEAT~\citep{stanley2002evolving} and generates the substrate network: regions where CPPN output variance exceeds a threshold are subdivided, yielding ${\sim}$10 hidden neurons for 2-input Boolean tasks.

Indirect encoding is used because specifying network topologies with heterogeneous per-node activations by hand is intractable: the combinatorial space grows exponentially with network size.

For each task, the same network is evaluated with a task-specific NT vector and activation function (Algorithm~\ref{alg:eval}, Figure~\ref{fig:architecture}); there are no separate output pathways. Fitness is the product of per-task accuracies: $F = \prod_t a_t$, enforcing simultaneous mastery. The intentionally small, single-layer substrate isolates the multi-task barrier from confounding capacity effects.

\subsection{Boolean Logic Benchmark}

We use five Boolean tasks: XOR (non-linearly separable), AND and OR (linearly separable with different thresholds), and NAND and NOR (complements). These span two computational classes, parity (XOR) and threshold detection (the rest), providing a minimal test of activation function diversity. Boolean tasks offer unambiguous ground truth and isolate the activation-modulation interaction from confounds (noise, dimensionality, function approximation error).

\subsection{Activation Functions}

The baseline condition applies a uniform activation function (tanh) to all hidden neurons across all tasks. The per-task condition assigns sin to parity tasks (XOR) and tanh to threshold tasks (AND, OR, NAND, NOR). Robustness experiments extend the palette to 10 functions spanning two classes: \emph{monotonic} (tanh, sigmoid, ReLU, leaky ReLU, identity) and \emph{oscillatory} (sin, cos, sin(2x), GCU~\citep{noel2021gcu}, dampened sin).

\subsection{Experimental Design}

A run \emph{converges} when all tasks reach $\geq$98\% accuracy; the convergence rate (fraction of seeds converging) is the default success metric. Convergence rates are compared with Fisher's exact test (two groups) and $\chi^2$ (multiple groups), with Cram\'{e}r's V ($\hat{\phi}$) as the effect size (association strength, 0--1). Unless noted, \emph{significant} means $p < 0.01$ ($\alpha = 0.01$) after Holm-Bonferroni correction; exact $p$-values appear in Table~\ref{tab:key-results}, and non-significant results are noted explicitly.

We ran more than 10,000 experiments (Table~\ref{tab:experiments}): 3,969 baseline experiments systematically varying task combinations, topologies, population sizes, and seeds; plus 6,570 robustness experiments across 46 conditions. All experiments use JAX~\citep{bradbury2018jax} on CPU. Substrate parameters: max resolution depth 4--5, variance threshold 0.03. NEAT: compatibility threshold 2.5, max stagnation 40. Neuromodulation strength $s = 5.0$, gains clipped to $[0.1, 5.0]$. Unless otherwise noted, every reported condition uses 30 independent random seeds (replications); seed counts are stated only where they differ from 30.

\begin{table*}[t]
\centering\small
\caption{Experiment summary (3,969 baseline + 6,570 robustness = 10,539 total).}
\label{tab:experiments}
\begin{tabular}{@{}lrc|lrc@{}}
\toprule
\textbf{Baseline} & \textbf{Runs} & \textbf{Gens} & \textbf{Robustness} & \textbf{Runs} & \textbf{Gens} \\
\midrule
Benchmark (combos $\times$ topo $\times$ pop $\times$ seeds) & 1,665 & 100 & Optimizer independence (Adam vs.\ ES) & 120 & 100/2K \\
Validation + ablation (per-task, 2-task, comp.) & 2,304 & 100--200 & Oscillatory class (cos, sin2x, GCU, damp.\ sin) & 300 & 100--200 \\
& & & Higher arity (3- through 5-input) & 960 & 200 \\
& & & Multi-layer depth (1--5 layers, 5--10 tasks) & 780 & 100--200 \\
& & & Population \& parameter sensitivity & 930 & 100--200 \\
& & & Domain generality (2D, 4D, Gaussian XOR) & 1,200 & 200 \\
& & & Task scaling + NT geometry & 2,040 & 100--200 \\
& & & Monotonic activation ablation (5 act.\ $\times$ 30) & 150 & 100 \\
& & & Adam optimizer 3-input (2 cond.\ $\times$ 30) & 60 & 2K \\
& & & Extended generations (10$\times$, indirect 5-task) & 30 & 1K \\
\midrule
\textbf{Baseline total} & \textbf{3,969} & & \textbf{Robustness total} & \textbf{6,570} & \\
\bottomrule
\multicolumn{6}{r}{\textbf{Grand total: 10,539}} \\
\end{tabular}
\end{table*}

\begin{algorithm}[t]
\caption{Multi-task evaluation of a single network}
\label{alg:eval}
\begin{algorithmic}[1]
\REQUIRE Network weights $\mathbf{W}$, receptor densities $\mathbf{R}$, base gains $\mathbf{g}_{\text{base}}$, task set $\mathcal{T}$
\ENSURE Product fitness $F$
\STATE $F \leftarrow 1.0$
\FOR{each task $t \in \mathcal{T}$}
    \STATE Look up NT vector $\mathbf{NT}_t$ and activation $f_t$
    \STATE Modulation: $\mathbf{p} = \mathbf{R}_{:,1{:}3}\, \mathbf{NT}_{t,1{:}3}$
    \STATE Forward: $\mathbf{h}_{\text{out}} {=} f_t(\mathbf{g}_{\text{eff}} {\odot}\, \mathbf{h} {+} \mathbf{b}_{\text{mod}}) {\odot}\, \boldsymbol{\gamma}$
    \IF{$\text{ACh}_t = 0.0$}
        \STATE Invert output: $y \leftarrow 1 - y$ \hfill (Eq.~\ref{eq:ach})
    \ENDIF
    \STATE Evaluate accuracy $a_t$ on all input patterns
    \STATE $F \leftarrow F \times a_t$
\ENDFOR
\RETURN $F$
\end{algorithmic}
\end{algorithm}

\section{Results}
\label{sec:results}

\subsection{The Multi-Task Barrier}

With uniform tanh activation, the success rate declines from 100\% (1-task) through 68\% (2-task) and 34\% (3-task) to 0\% (5-task) across all 54 such experiments. Networks consistently reach 100\% on threshold tasks but plateau at exactly 75\% on XOR (3 of 4 input patterns correct). A subsequent replication (630 runs) confirms: all XOR-free conditions converge (100\%), while XOR collapses to 0\% whenever combined with AND or NAND. Extending the evolutionary budget tenfold, to 1,000 generations, does not break the ceiling (0\%, XOR pinned at 0.750), confirming a convergence barrier rather than insufficient runtime. The 2-task benchmark is pair-specific: AND+OR achieves 100\% while XOR+AND achieves 0\% (Table~\ref{tab:2task}).

\subsection{The 75\% Ceiling: Monotonicity vs.\ XOR}
\label{sec:diagnosis}

The \emph{zero-variance} nature of the 75\% plateau is the key diagnostic. A \emph{monotonic} activation can only increase (or only decrease); once it maps a higher input to HIGH, all higher inputs also map HIGH. An \emph{oscillatory} activation changes direction, enabling $f(x_1) < f(x_2) > f(x_3)$ for $x_1 < x_2 < x_3$.

\begin{figure}[h!]
\centering
\includegraphics[width=\columnwidth]{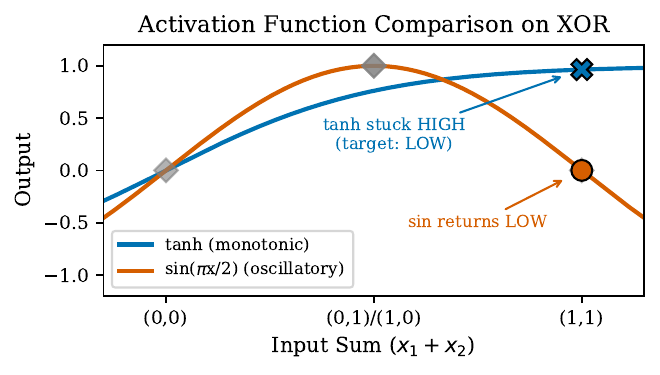}
\caption{Activation function comparison on XOR (output versus input sum $x_1+x_2$). The monotonic tanh (blue) stays HIGH at the largest sum, where XOR needs LOW, scoring 3/4 ($=$ 75\%); the oscillatory sin (orange) returns LOW there, solving XOR (4/4 $=$ 100\%).}
\label{fig:results_overview}
\end{figure}

The intuition (Figure~\ref{fig:results_overview}): each tanh neuron is monotonic in its weighted input sum, so once it maps sum$=$1 to HIGH, it cannot return LOW for sum$=$2. While multiple hidden neurons with appropriate output weights \emph{can} combine to produce non-monotonic mappings (as confirmed by the optimizer independence test below), evolutionary search consistently fails to discover the tightly coupled weight-receptor configuration that solves XOR jointly with a conflicting task. Oscillatory activations bypass this: $\sin(\pi x/2)$ oscillates naturally (LOW$\to$HIGH$\to$LOW), matching XOR's alternating pattern at the single-neuron level.

\subsection{The Optimizer Independence Test}
\label{sec:optimizer}

To determine whether the barrier is representational (structurally impossible) or optimization-related, we test the \emph{same} neuromodulated single-output architecture (2$\to$10$\to$1) with Adam gradient descent~\citep{kingma2014adam} instead of evolutionary search (4 conditions):

\begin{itemize}[nosep]
    \item Adam + uniform tanh (lr$=$0.01): 100\%, median 190 steps
    \item Adam + uniform tanh (lr$=$0.001): 90\%
    \item Adam + per-task activation (lr$=$0.01): 100\%, median 40 steps ($4.75\times$ faster)
    \item $(\mu{+}\lambda)$-ES~\citep{beyer2002evolution} + uniform tanh: 76.7\%
\end{itemize}

The activation barrier is an evolutionary search barrier, not a representational limitation. The architecture \emph{can} represent all five tasks with uniform tanh. Adam's gradients navigate the coupled weight-receptor space that the $(\mu{+}\lambda)$-ES we tested did not. Per-task activation remains beneficial even with Adam ($4.75\times$ faster). For evolutionary systems, the primary context of open-ended artificial life, oscillatory activations provide the evolutionary shortcut that makes multi-task neuromodulation feasible without gradient information.

\subsection{Per-Task Activation Functions}
\label{sec:breakthrough}

We test per-task hidden-layer activation: sin for XOR, tanh for threshold tasks. Weights, receptor densities, and the output neuron remain shared. Only the activation and NT vector change between task evaluations. This is analogous to how biological circuits use neurons with different input-output properties for different functional roles~\citep{zeng2017neuronal}.

\begin{figure}[h!]
\centering
\includegraphics[width=\columnwidth]{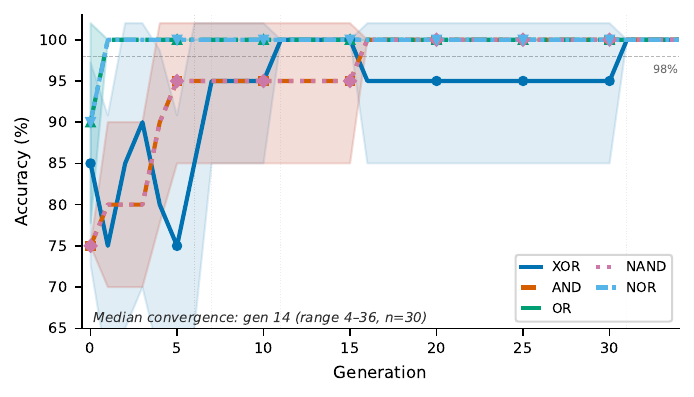}
\caption{100\% simultaneous 5-task success: per-task accuracy across generations. XOR (sin) breaks the 75\% ceiling and converges last.}
\label{fig:pertask_hero}
\end{figure}

Validated (Pop$=$750, product aggregation): all seeds achieve 100\% on all five tasks simultaneously. Median convergence: 14 generations (95\% CI: [12.7, 19.2]; mean $15.9 \pm 8.8$, range 4--36). Only OR and NOR converge by generation 1; AND and NAND follow, with XOR converging last, requiring additional generations to tune sin's oscillatory phase (Figure~\ref{fig:pertask_hero}).

\subsection{Ablation Studies}
\label{sec:ablations}

\textbf{Neuromodulation is necessary.} Per-task activation with \emph{flat} NT vectors (identical $[0.5, 0.5, 0.5, 1.0]$ for all tasks): none converge. XOR achieves 100\% (sin suffices alone) but all four threshold tasks plateau at exactly 50\% (chance). Without task-specific modulation, the network cannot differentiate AND (decision threshold $\theta \approx 1.5$) from OR ($\theta \approx 0.5$). Removing gain/bias modulation while preserving inversion (flattening DA/5HT/NE but keeping correct ACh) yields no convergence, stuck at the 75\% ceiling.

\textbf{Uniform sin suffices on this benchmark.} Uniform sin on all five tasks: all seeds converge (median 17, range 2--32). Sin can solve threshold tasks because its monotonically increasing region approximates a sigmoid. This confirms the barrier is specifically \emph{monotonic} activation: tanh fails because it cannot oscillate, not because a single type is insufficient. Per-task assignment converges no faster here: on the 25 seeds the two conditions share, the difference is not significant (median 14 vs.\ 17 overall, 15 vs.\ 17 seed-matched; Wilcoxon signed-rank $p = 0.88$, paired mean difference $-0.1$ generations). Its advantage is \emph{design clarity} rather than speed: matching primitives to task structure avoids relying on incidental properties that may not generalize. That distinction is not cosmetic: the sufficiency of uniform sin holds for this 2-input Boolean set and does not transfer, reversing on continuous regression, where uniform sin drops to 6.7\% and uniform tanh reaches 100\% (\nameref{sec:robustness}).

\textbf{Per-task activation is the critical component.} Removing per-task activation (uniform tanh with neuromodulation): 0\% success. Removing neuromodulation (per-task with flat NT): 0\% success. Neither suffices alone. Together they enable 100\%.

\textbf{Failed approaches.} Per-task receptor densities (evolving separate $\mathbf{R}$ per task) expanded the search space from $N{\times}4$ to $N{\times}4{\times}5$ parameters, yielding 50\% fitness. Output blending (mixing task outputs instead of evaluating separately) masked XOR's failure by averaging across tasks, settling at the 75\% ceiling. Sequential evaluation (one task at a time, switching NT) succeeds under per-task activation, confirming the barrier is activation-task mismatch, not the evaluation protocol.

\textbf{Substrate reuse.} Each validated network is a single evolved substrate that reconfigures its behavior at inference: the same weights and receptor densities solve all five Boolean tasks by switching the NT vector and activation function.

\textbf{Formulation ablation.} Each neuromodulation component (gain, bias, gating) is independently sufficient at 2-input: all three achieve 100\% when tested in isolation, though convergence is $1.7\times$ slower than the combined formulation. At 3-input, this robustness breaks down: bias-only remains fully sufficient (100\%), but gain-only degrades substantially to 63.3\%, significantly lower than the full formulation. Bias modulation provides the task-specific offsets for differentiating 8-pattern truth tables at higher arity.

\textbf{2-task compatibility.} Per-task activation eliminates all 2-task barriers (Table~\ref{tab:2task}: all 10 pairs converge). The XOR+AND pair (0\% under both uniform conditions) illustrates the core mechanism: these tasks require incompatible hidden-layer computations under monotonic activation, a conflict that per-task activation resolves.

\begin{table}[t]
\centering\small\setlength{\tabcolsep}{2pt}
\caption{2-task convergence rates ($\geq$98\%, n$=$30 per pair). \emph{Independent}: each task has a unique 4D NT vector (ACh$=$1.0 for all). \emph{Compositional}: NAND/NOR reuse AND/OR profiles with ACh$=$0.0 for output inversion (Eq.~\ref{eq:ach}). Per-task activation eliminates all barriers.}
\label{tab:2task}
\begin{tabular}{@{}lccc@{}}
\toprule
\textbf{Pair} & \textbf{Indep.\ uniform} & \textbf{Comp.\ uniform} & \textbf{Comp.\ per-task} \\
\midrule
\multicolumn{4}{@{}l}{\emph{XOR-containing pairs}} \\
XOR+NOR  & 100\% & 83.3\% & 100\% \\
XOR+OR   & 86.7\% & 86.7\% & 100\% \\
XOR+NAND & 46.7\% & 0\%   & 100\% \\
XOR+AND  & 0\%    & 0\%   & 100\% \\
\midrule
\multicolumn{4}{@{}l}{\emph{Threshold-only pairs}} \\
AND+OR   & 100\% & 100\% & 100\% \\
AND+NAND & 100\% & 100\% & 100\% \\
OR+NAND  & 100\% & 100\% & 100\% \\
NAND+NOR & 100\% & 100\% & 100\% \\
OR+NOR   & 0\%   & 100\% & 100\% \\
AND+NOR  & 16.7\% & 100\% & 100\% \\
\bottomrule
\end{tabular}
\end{table}

\subsection{Oscillatory Class Generalization}

The following experiments use direct-encoded controls to isolate the activation-modulation interaction from CPPN-specific effects (the main results appear above). We evaluate four oscillatory activations plus a tanh control under the same neuromodulated architecture (5-task, compositional NT design, direct-encoded $(\mu{+}\lambda)$-ES). All four oscillatory functions achieve 100\%: cos (median 4 gen), sin(2x) (median 3.5), GCU (median 3), dampened sin (median 4.5). The tanh control achieves only 76.7\%, with all 7 failures showing XOR stuck at 75\%. Oscillatory functions converge roughly 4--6$\times$ faster than tanh. A separate monotonic ablation (150 runs) confirms the complementary pattern: identity 0\%, leaky ReLU 20\%, sigmoid 33.3\%, ReLU 40\%, tanh 76.7\%. Every failure shows XOR at exactly 0.750. The divide is qualitative: oscillatory functions eliminate the barrier entirely, while no monotonic function reaches 100\%.

\textbf{Arity-dependent class advantage.} At 3-input, the oscillatory class advantage persists but is no longer uniform (150 runs, 5 conditions): cos 90\%, GCU 90\%, dampened sin 56.7\%, sin(2x) 40\%, tanh 26.7\%. Standard-period functions (cos, GCU) significantly outperform tanh, while sin(2x) does not. The doubled frequency creates more zero-crossings, producing a rugged fitness surface comparable to monotonic activations at higher arity.

\subsection{Robustness}
\label{sec:robustness}

Forty-six robustness conditions (6,570 runs; Table~\ref{tab:experiments}) probe parameter sensitivity, depth, arity, topology, encoding, and domain generality.

\textbf{Parameter sensitivity.} Sweeping modulation strength $s$ from 0.5 to 20.0 (180 runs): $s = 1.0$ through $s = 20.0$ all achieve 100\%. Only $s = 0.5$ shows slight degradation (93.3\%). Sweeping population from 100 to 1,000 (150 runs): all conditions achieve 100\% convergence.

\textbf{Multi-layer depth alternative.} Direct-encoded MLPs with neuromodulation under uniform tanh: 1-layer 76.7\%, 2-layer 100\%, 3-layer 100\%. Per-task activation at 1 layer remains $4\times$ faster (median 3 vs.\ 12 gen) with $2.9\times$ fewer parameters.

\textbf{Higher arity.} The barrier extends to higher-arity tasks: 1-layer tanh achieves 0\% on Parity-4 (4-input). At 3-input, per-task activation achieves 100\% (5-task) across all tested populations (Pop $\in\{100, 300, 1500\}$), uniform tanh 26.7\%. However, uniform tanh at 3-input exhibits a \emph{search-budget ceiling}: Pop$=$1500 achieves 50\%, with no further improvement at Pop$=$2000--3000. An optimizer independence test at 3-input confirms this is an evolutionary search barrier, not a structural limit: Adam + uniform tanh achieves 100\% (median 130 steps), bypassing the ceiling entirely. At 4-input with Pop$=$1500: per-task improves marginally but not significantly to 30\% (from 13.3\% at Pop$=$750), while 2-layer tanh reaches 73.3\%.

\textbf{Capacity and depth interact.} Doubling hidden neurons from 20 to 40 within one layer: 0\% with tanh at both sizes. The barrier is not capacity-limited. However, capacity helps \emph{with depth}: 2-layer tanh at 40 neurons achieves 76.7\% vs.\ 43.3\% at 20 neurons, with Parity-4 solved in all seeds. The bottleneck shifts entirely to AND-4 and OR-4.

\textbf{Topology neutrality is arity-dependent.} At 2-input, all six topologies (feedforward, hidden-only recurrent, backward, lateral, self-connections, and full recurrent) produce identical per-seed convergence. At 3-input with uniform tanh, feedforward achieves 20\% while full recurrent achieves 80\% (significant). Per-task activation saturates this effect: both feedforward and full recurrent achieve 100\% at 3-input (vs 20\%/80\% under uniform tanh).

\textbf{IMPLY barrier.} The barrier extends beyond parity to asymmetric tasks. IMPLY ($[1,1,0,1]$) requires distinguishing $(0,1){\to}1$ from $(1,0){\to}0$, a directional relationship that symmetric gain/bias/gating modulation cannot encode. Sin resolves it (median 10 gen), while ReLU achieves only 6.7\% and ELU 0\%.

\textbf{Task scaling.} Adding XNOR (the sixth Boolean gate, sharing XOR's hidden-layer modulation with ACh$=$0.0) yields 100\% at 6 tasks. Compositionally related tasks add no modulatory burden. The cliff is sharp: 7, 8, and 10 tasks yield 0\% under indirect encoding. IMPLY accounts for all failures. Multi-layer depth extends the limit: 2-layer achieves 100\% at 8 tasks (both uniform tanh and per-task), 80\% at 9 tasks, and 30--50\% at 10 tasks. At 10 tasks, activation diversity actively \emph{hurts}: 10 unique activations yield 0\%, while simplified per-task (sin for parity, tanh for rest) achieves 30\% and uniform tanh 43.3\%. The implication-family tasks (IMPLY, NIMPLY, CONVERSE, CONVERSE-NIMPLY), each a directional, asymmetric function, are the sole failure points at 10 tasks.

\textbf{Frozen substrate.} Weights trained under tanh on threshold tasks produce chance-level XOR (50\%) even when sin activation is applied post-hoc (0\% XOR solved), while all threshold tasks remain at 100\%. Retraining from these frozen weights with per-task activation converges in 2 generations. The barrier operates during training dynamics, not at evaluation.

\textbf{5-input extension.} At 5-input, per-task achieves 83.3\% vs.\ 2-layer 60\%, while uniform tanh remains at 0\%. Per-task activation is the most parameter-efficient path (276 vs.\ 1,026 parameters).

\textbf{Continuous domain.} The barrier is domain-specific, and on smooth regression it reverses. Continuous regression (three tasks, n$=$30 per condition): uniform tanh 100\%, per-task 43.3\%, uniform sin 6.7\%. This is the only condition in which uniform tanh outperforms uniform sin. Where no task carries parity structure, monotonicity is the asset and oscillation the liability, which is the same principle that makes sin necessary on the Boolean benchmark read in the opposite direction. However, Gaussian XOR (continuous with parity-class structure): tanh 3.3\% vs.\ per-task 50.0\% vs.\ 2-layer tanh 93.3\%. Multi-class discrete classification (4-class Gaussian XOR): tanh 30\% vs.\ per-task 100\%. Higher-dimensional continuous domains (4D Iris-like synthetic): only 2-layer succeeds (80\%). The activation-task interaction depends on task structure, not domain type.

\textbf{Per-task activation versus uniform sin.} The two conditions tie on the 2-input Boolean benchmark, but not beyond it. Fifteen robustness conditions ran both arms under matched settings and identical seeds (900 runs). Per-task activation converges more often in 11 of them and less often in none; the remaining four are ties in which neither condition ever converges. Because the arms are seed-matched, the comparison is paired: per-task converged on 75 runs where uniform sin failed, against 5 in the opposite direction (exact McNemar, $p < 10^{-15}$; sign test over conditions, $p = 0.001$). Pooled across the fifteen, per-task reaches 42.4\% versus 26.9\%. The margin is widest where the task set mixes computational classes: Parity-3 multi-task 100\% vs.\ 70\%, 5-task 3-input 100\% vs.\ 80\%, Gaussian XOR 50\% vs.\ 6.7\%. Matching activations to task structure therefore buys robustness rather than speed, which is why uniform sin's sufficiency on one benchmark does not make it the safer default.

\begin{table*}[t]
\centering
\caption{Summary of key results. ``Repl.'' = number of replications; ``Rate'' = convergence rate (all tasks $\geq$98\%). $p$-values: Fisher's exact test. $\hat{\phi}$: Cram\'{e}r's V.}
\label{tab:key-results}
\small
\begin{tabular}{@{}llccclc@{}}
\toprule
\textbf{Finding} & \textbf{Condition} & \textbf{Repl.} & \textbf{Rate} & \textbf{vs.\ Control} & \textbf{$p$-value} & \textbf{$\hat{\phi}$} \\
\midrule
5-task barrier & Uniform tanh, 1L, ES & 54 & 0\% & --- & --- & --- \\
Per-task activation & Sin+tanh, 1L, ES & 30 & 100\% & vs.\ 0\% tanh & $<10^{-10}$ & 1.00 \\
Adam bypasses barrier & Uniform tanh, 1L, Adam & 30 & 100\% & vs.\ 76.7\% ES & 0.011 & 0.36 \\
Oscillatory class & cos/sin2x/GCU/damp.\ sin & 120 & 100\% & vs.\ 76.7\% tanh & $<10^{-5}$ & 0.44 \\
Capacity irrelevant & 40 neurons, tanh, 1L & 30 & 0\% & vs.\ 0\% (20n) & 1.000 & 0.00 \\
Neuromod necessary & Flat NT + per-task act. & 30 & 0\% & vs.\ 100\% full & $<10^{-10}$ & 1.00 \\
Depth resolves & Uniform tanh, 2L & 30 & 100\% & vs.\ 76.7\% (1L) & 0.011 & 0.36 \\
IMPLY barrier & Sin for IMPLY, 7-task & 30 & 100\% & vs.\ $\leq$6.7\% mono. & $<10^{-10}$ & 1.00 \\
Frozen substrate & Tanh-trained + sin eval & 30 & 0\% XOR & vs.\ 100\% fresh & $<10^{-10}$ & 1.00 \\
3-input extends & Per-task, 5-task, 3-input & 30 & 100\% & vs.\ 26.7\% tanh & $<10^{-5}$ & 0.76 \\
\bottomrule
\end{tabular}
\end{table*}

\subsection{Task Compatibility in Modulatory Space}
\label{sec:compatibility}

What determines whether two tasks can coexist? Euclidean NT distance alone is a poor predictor of convergence rate ($R^2 = 0.31$ across 20 pair$\times$schema observations). The clearest dissociation is OR+NAND vs.\ OR+NOR: \emph{identical} Euclidean distance ($d = 0.566$) but 100\% vs.\ 0\%. The difference is truth table conflict (2/4 disagreements vs.\ 4/4). Compatibility depends on the interaction of four factors: (1)~\emph{truth table conflict}: pairs with $\leq$2/4 disagreements achieve 100\% regardless of distance; (2)~\emph{NT distance}: for complementary pairs (4/4 conflict), distance predicts convergence monotonically (AND+NAND at $d = 1.200$: 100\%; AND+NOR at $d = 0.894$: 16.7\%; OR+NOR at $d = 0.566$: 0\%); (3)~\emph{NT direction}: reorienting OR+NOR via mirror reflection (swapping DA$\leftrightarrow$NE) at identical distance restores 100\% at generation 1 because the baseline direction is degenerate for differential gain modulation; (4)~\emph{output polarity}: toggling NOR's ACh from 1.0 to 0.0 at identical distance restores 100\% at generation 0, because inversion makes complementary truth tables trivially compatible via output sign flip. Three independent mechanisms can rescue OR+NOR: polarity inversion (ACh), direction reorientation, or sufficient distance increase. All compatibility patterns vanish under per-task activation, confirming they are artifacts of activation-task mismatch.

\section{Discussion}
\label{sec:discussion}

\textbf{The barrier is evolutionary, not representational.} The optimizer independence test is the critical piece of evidence: the same architecture with uniform tanh achieves 100\% under Adam but 0\% under evolutionary search (indirect encoding) or 76.7\% (direct encoding). The frozen substrate experiment provides causal confirmation: tanh-trained weights produce chance-level XOR (50\%) even with sin applied post-hoc, proving the barrier operates during training dynamics, not at evaluation. The mutation-driven evolutionary search we tested did not discover the tightly coupled weight-receptor configuration within our budget, but gradients reliably do. For evolutionary systems, the context of open-ended artificial life, oscillatory activations are thus less a theoretical preference than a practical enabler. In embodied settings other mechanisms (an agent reorienting its body) could transform the same task, so this enabler is specific to the disembodied Boolean setting we study.

\textbf{Neuromodulation and activation function diversity are jointly necessary.} Neither mechanism suffices alone. Neuromodulation cannot make a monotonic function oscillate; per-task activation without modulation cannot shift decision boundaries between AND ($\theta \approx 1.5$) and OR ($\theta \approx 0.5$). Together, neuromodulation tunes the shared weight matrix while per-task activation provides the appropriate computational primitive~\citep{zeng2017neuronal,kepecs2014interneuron}. A richer neuromodulation that selects the activation function itself, not only its gain and bias, could unify the two mechanisms within a single modulatory signal.

\textbf{Alternative: multi-layer depth.} Two-layer tanh also reaches 100\%, but per-task activation at 1 layer is faster with fewer parameters. Depth eases the evolutionary search rather than adding representational capacity; a single layer already represents the solution (\nameref{sec:optimizer}).

\textbf{One genotype, many phenotypes.} Although the paper centers on the barrier, the result that matters for artificial life is what overcoming it enables (Figure~\ref{fig:architecture}): a single indirectly encoded genotype that expresses qualitatively different phenotypic behaviors depending on the task environment. This is environment-dependent phenotypic expression at the network level, a property relevant to open-ended systems where the task set is not known in advance.


\textbf{Implications for open-ended systems.} In evolutionary systems where gradient information is unavailable, activation-task mismatch creates barriers that standard mutation-driven search did not overcome within our budget. Three paths forward: (1)~\emph{evolved diversity}: include activation selection in the genotype, as shown in prior work on per-node function evolution~\citep{claret2026activations}, where evolution can meta-learn which functions to make available~\citep{claret2026bio}; (2)~\emph{developmental diversity}: evolve rules that differentiate neuron types during network formation~\citep{doursat2013review}; (3)~\emph{modular composition}: evolve specialized subnetworks with task-appropriate primitives, composed via routing. The 2-task compatibility matrix defines a space of achievable behavioral combinations that quality-diversity algorithms~\citep{pugh2016quality,mouret2015illuminating} could exploit.

\subsection{Limitations}

\begin{enumerate}[itemsep=2pt,parsep=0pt,topsep=2pt]
    \item \emph{Task complexity}: Boolean tasks provide clean analysis but are trivial. Higher-arity experiments confirm the barrier persists (Parity-4: 0\% tanh) and that per-task activation degrades at 4-input (13.3\%) though it remains fully effective at 3-input (100\%). Among the continuous domains investigated, those without parity-class structure show no activation barrier.
    \item \emph{Fixed NT presets}: NT vectors are manually specified; random NT experiments show one of ten sets achieves 100\% when ACh polarity is preserved, and co-evolving the NT vectors with substrate topology is a natural next step.
    \item \emph{Task identity}: Evaluation requires explicit task identity at test time. Biological neuromodulatory systems use context-dependent release~\citep{doya2002metalearning}; an artificial analog could infer task context from the input stream rather than receive it explicitly, though not necessarily through a separately learned module.
    \item \emph{Encoding interaction}: Indirect (CPPN) and direct encoding show different barrier strengths (0\% vs.\ 76.7\% at 1-layer tanh), though per-task activation resolves the barrier under both. The two encodings span genotype spaces of different size and structure, so this comparison is illustrative rather than strictly controlled.
    \item \emph{Evolutionary algorithm}: results are specific to $(\mu{+}\lambda)$-ES under direct and indirect encoding; multi-objective, novelty- and diversity-promoting, and quality-diversity search were not tested and could navigate the coupled weight-receptor space differently.
\end{enumerate}

\section{Conclusion}
\label{sec:conclusion}

Across more than 10,000 experiments, we show that monotonic activation functions create a systematic evolutionary search barrier in neuromodulated multi-task networks, imposing a 75\% ceiling on parity and asymmetric tasks that gradient-based optimization bypasses entirely. Per-task activation functions, matching oscillatory primitives to parity tasks and monotonic to threshold tasks, eliminate the barrier completely: 100\% 5-task success (median 14 generations). The oscillatory class generalizes beyond sin (cos, sin(2x), GCU, dampened sin), and the result is robust across a 40$\times$ range of modulation strengths.

Three findings should change practice in evolutionary multi-task systems: (1)~audit the activation palette against the task set; (2)~for tasks spanning computational classes, prefer oscillatory activations (uniform sin: 100\% vs.\ tanh: 0\%); (3)~exploit output-level reuse via polarity inversion (Eq.~\ref{eq:ach}). To the question posed in the abstract: yes, when the activation palette matches the task set's computational demands, indirect encoding with neuromodulation lets evolution discover multi-behavioral substrates where one evolved topology expresses different behaviors across tasks. Computational primitives should be treated as an evolvable trait, not a fixed design choice.


\section{Acknowledgments}
Generative AI tools assisted with visualizations, pseudocode conversion, and draft revision.

\footnotesize
\bibliographystyle{apalike}
\bibliography{references}

\end{document}